\documentclass{acm_proc_article-sp}
\usepackage{subcaption}
\usepackage{hyperref}
\usepackage[ruled]{algorithm2e}

\begin{document}
\CopyrightYear{2015} 
\crdata{978-1-4503-2431-1/13/11} 
\clubpenalty=10000 
\widowpenalty = 10000

\title{Sensor-Type Classification in Buildings
}
\numberofauthors{1} 
\author{
\alignauthor
Dezhi Hong$^1$, Jorge Ortiz$^2$, Arka Bhattacharya$^3$, Kamin Whitehouse$^1$\\
	\affaddr{$^1$Department of Computer Science, University of Virginia, VA USA}\\
	\affaddr{$^2$IBM T.J. Watson Research Center, Yorktown Heights, NY USA}\\
	\affaddr{$^3$Computer Science Division, University of California Berkeley, CA USA}\\
	\email{\{hong, whitehouse\}@virginia.edu, jjortiz@us.ibm.com, arka@eecs.berkeley.edu}
}

\maketitle
\begin{abstract}
Many sensors/meters are deployed in commercial buildings to monitor and optimize their performance.  
However, because sensor metadata is inconsistent across buildings, software-based
solutions are \emph{tightly coupled} to the sensor metadata conventions (i.e. schemas and naming) for each building. 
Running the same software across buildings requires significant integration effort.


Metadata normalization is critical for scaling the deployment process and allows us to \emph{decouple} building-specific conventions 
from the code written for building applications.  It also allows us to deal with \emph{missing} metadata.  One important aspect of 
normalization is to differentiate sensors by the \emph{type}
of phenomena being observed.
In this paper, we propose a general, simple, yet effective classification scheme to differentiate sensors in buildings by type.
We perform ensemble learning on data collected from over 2000 sensor streams in two buildings.
Our approach is able to achieve more than 92\% accuracy for classification within buildings and more than 82\% accuracy for across buildings. 
We also introduce a method for identifying potential misclassified streams.  This is important because it allows us to identify opportunities to 
attain more input from experts -- input that could help improve classification accuracy when ground truth is unavailable.  We show that
by adjusting a threshold value we are able to identify at least 30\% of the misclassified instances.
\end{abstract}


\section{Introduction}



Commercial buildings are sites of large sensor/meter deployments used to monitor and optimize their performance.  
With the recent interest in reducing building energy consumption and
increasing their efficiency, it
is important to consider ways to quickly bootstrap a set of building data streams
into an analytical pipeline, such as overall building efficiency or
comfort-assessment analytics and control.
However, because sensor metadata is inconsistent across buildings, software-based
solutions are \emph{tightly coupled} to the sensor metadata conventions (i.e. schemas and naming) for
each building. 
Running the same software across buildings requires significant integration effort.

Current `point' naming conventions and unsystematic recording of metadata form a 
bottleneck in deployment scalability for analytics jobs.  A `point' refers to a 
physical location where a sensor is taking measurements. 
Each building vendor uses their own naming scheme and
unique variants of each scheme are implemented from building to building; variations exist
even across buildings that have contracted the same vendor.
In addition, expanded descriptive information about the point is sometimes unavailable 
-- so determining their meaning is painfully slow or impossible.  
Because these are conventions carried
out by humans, they are inconsistent within and across building data sets.
This makes the integration process laborious for building experts and a non-starter for 
non experts.  The process is fundamentally unscalable.

Consider a simple analysis program, which has the ability
to identify anomalous readings from a specific kind of sensor. To execute this job, 
the process organizes each sensor by type and location, generates the distribution of
readings across them, and identifies broken sensors where some fraction of
their readings are above some threshold value on the distribution.
The identification step in the process is the most challenging 
because of the problems described.  Ideally, the program would search for points the
way you search for web pages in a search engine -- using semantically meaningful 
terminology. 

Point names contain set of codes that are 
semantically meaningful to the building manager of a specific building.
For example, the point \texttt{BLDA1R435\_\_ART} is constructed as a concatenation of such codes.
The name of the building (first 4 characters), the air handling unit identifier (the 
fifth character), the room number (R435), and the type ART (area room temperature) -- which 
indicates that this are measurement is produced by a temperature sensor. In addition
to point names, there may be some descriptive metadata.  The description for this point (if it 
exists) could describe that this is a ``temperature sensor in room 435''.
However, since point names do not follow the exact same structure within and across
buildings (and certainly do not follow the same convention across vendors)
no single approach could solve the normalization problem.  A suite of approaches is necessary.

Metadata normalization is critical for scaling the software deployment process.  
It allows us to
\emph{decouple} building-specific conventions from the code written for building applications.  Normalization allows us to
boost existing metadata, correct incorrect metadata, or generate common metadata when it is missing altogether.
One such component in the normalization suite should differentiate sensor feeds by \emph{type}.  For example, we should be able to 
differentiate between sensors measuring 
temperature from sensors measuring pressure.  In addition, we should be able to use what we learn from one building and apply
it to another.  This is especially useful in cases where similar stream types are labeled differently, labeled incorrectly,
 or not labeled at all.

Normalization would allow us to quickly run jobs across many sites by enabling wide \emph{searchability} of points across many buildings at once.
In order to meaningfully deal with disparate building streams in a scalable 
fashion the streams should be \emph{searchable} across various properties, such
as building name, room location, and type.  Moreover, we
assert that wide searchability is necessary for achieving scalability.  By providing a tool for
searching across building streams, we minimize the deployment time for applications;  
allowing them to be used in \emph{all} buildings, not just a single one.

One of the important aspect of the sensor meta/data that we can leverage are the actual patterns in the readings themselves.
Deep inspection of features in the data can yield meaningful results about the \emph{type} of data that it is
and can help us with the label normalization problem.  This paper examines this path using standard machine learning
approaches.  We observe that statistical features over small time windows can be used to identify the stream type.
Moreover, we show that the classification of stream-type can be achieved using an ensemble of classifiers which is
known to outperform a single classifier.

We conduct a comprehensive study on the data collected from over 2000 sensors in two separate buildings on two campuses. Our main contributions are:

\begin{itemize}
\item We propose a simple, general yet effective feature extraction scheme to achieve sensor type classification in the context of commercial buildings.
\item We formulate an approach to identifying potential misclassified sensor streams (in terms of the type classes) when no ground truth labels are available.
\item We evaluate our classification technique using data from over 2000 sensor series of 6 types in two buildings on two campuses, and our technique is able to achieve around 92\% and 98\% accuracy when doing classification within each building, and around 82\% accuracy when inferring type information across buildings.
\item We also evaluate our solution to misclassification identification and the results demonstrate that we are able to identify at least 30\% percent of the target population by choosing an optimal threshold for decision.
\end{itemize}

We believe this is an important study given the recent trends in the penetration
of the \emph{internet of things} into our homes and environments.  Studies show that normalization is an especially pernicious
and widely ubiquitous problem in embedded systems, with only 7\% of data tagged and only 1\% analyzed~\cite{kpcb}.
Our technique can be used to unify that data across many deployment and enable broad search 
and exploration of new applications.  For example, sensing device names for 
the internet of things are likely to follow similar conventions with very little 
context.  We argue that unification through boosting will be necessary in this broader domain.






\begin{figure*}[ht!]
\centering
  \begin{subfigure}{0.32\textwidth}
                \centering
    \includegraphics[width=\textwidth]{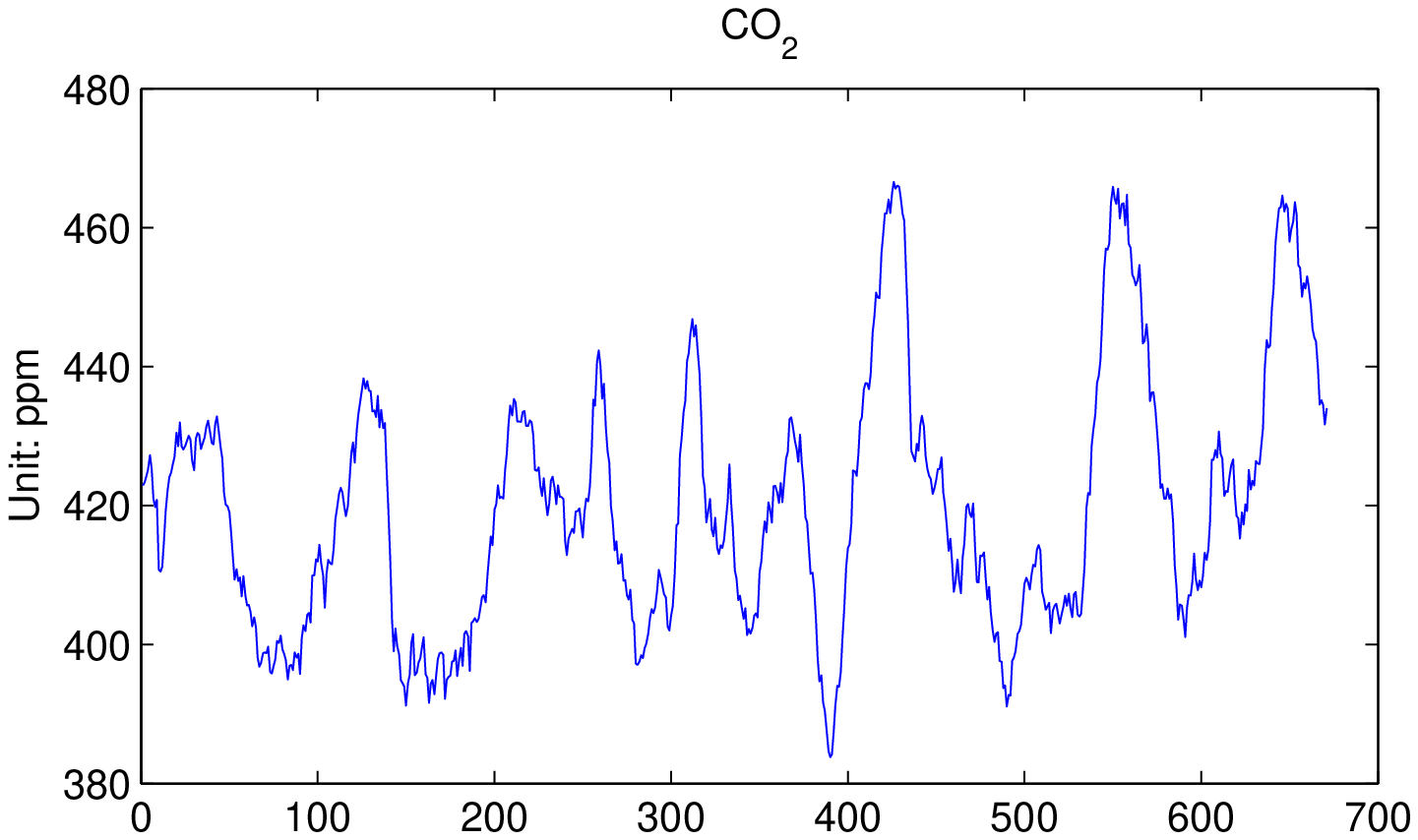}
                \caption{$CO_{2}$}
  \end{subfigure}
  \begin{subfigure}{0.32\textwidth}
                \centering
    \includegraphics[width=\textwidth]{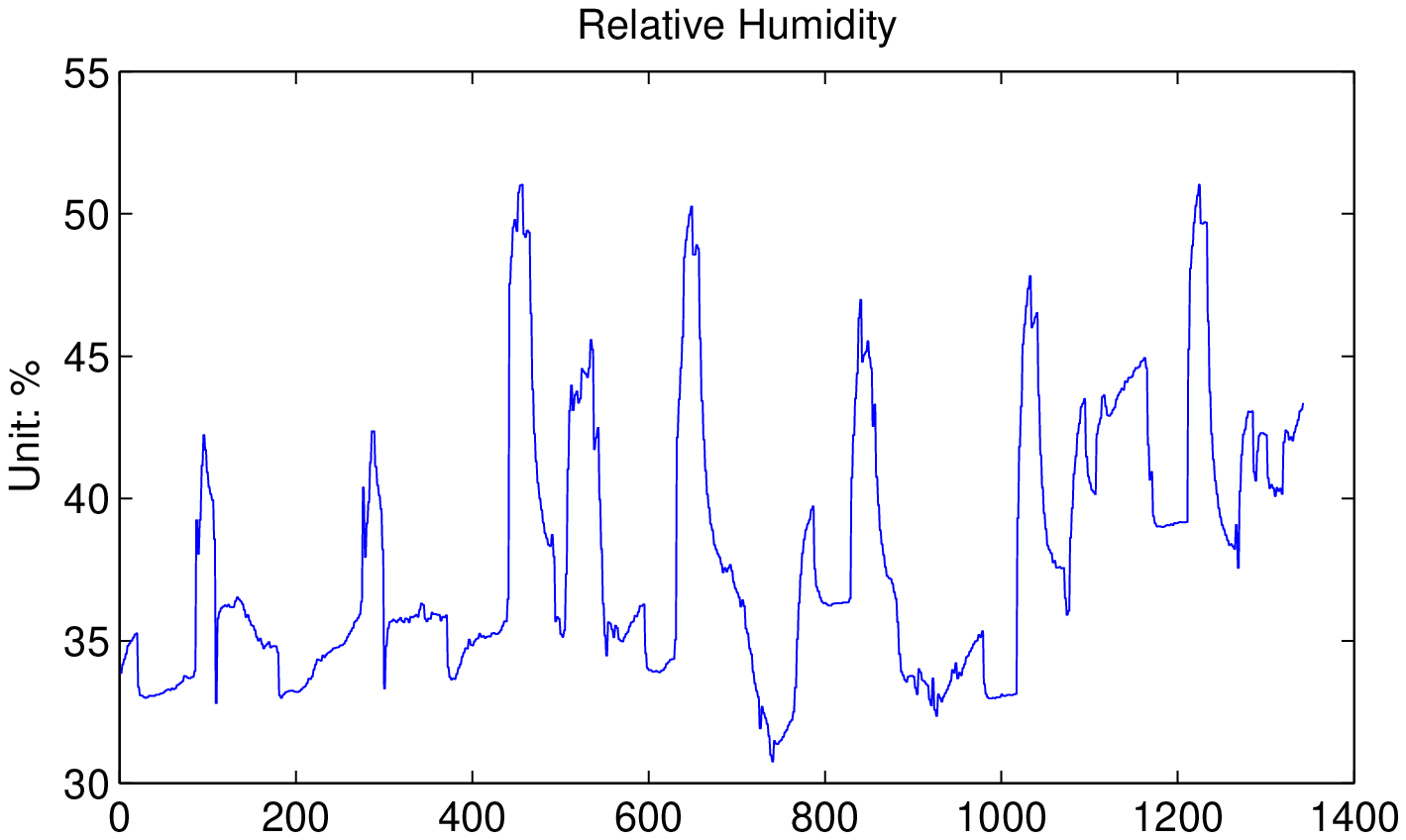}
                \caption{Humidity}
  \end{subfigure}
  \begin{subfigure}{0.32\textwidth}
                \centering
    \includegraphics[width=\textwidth]{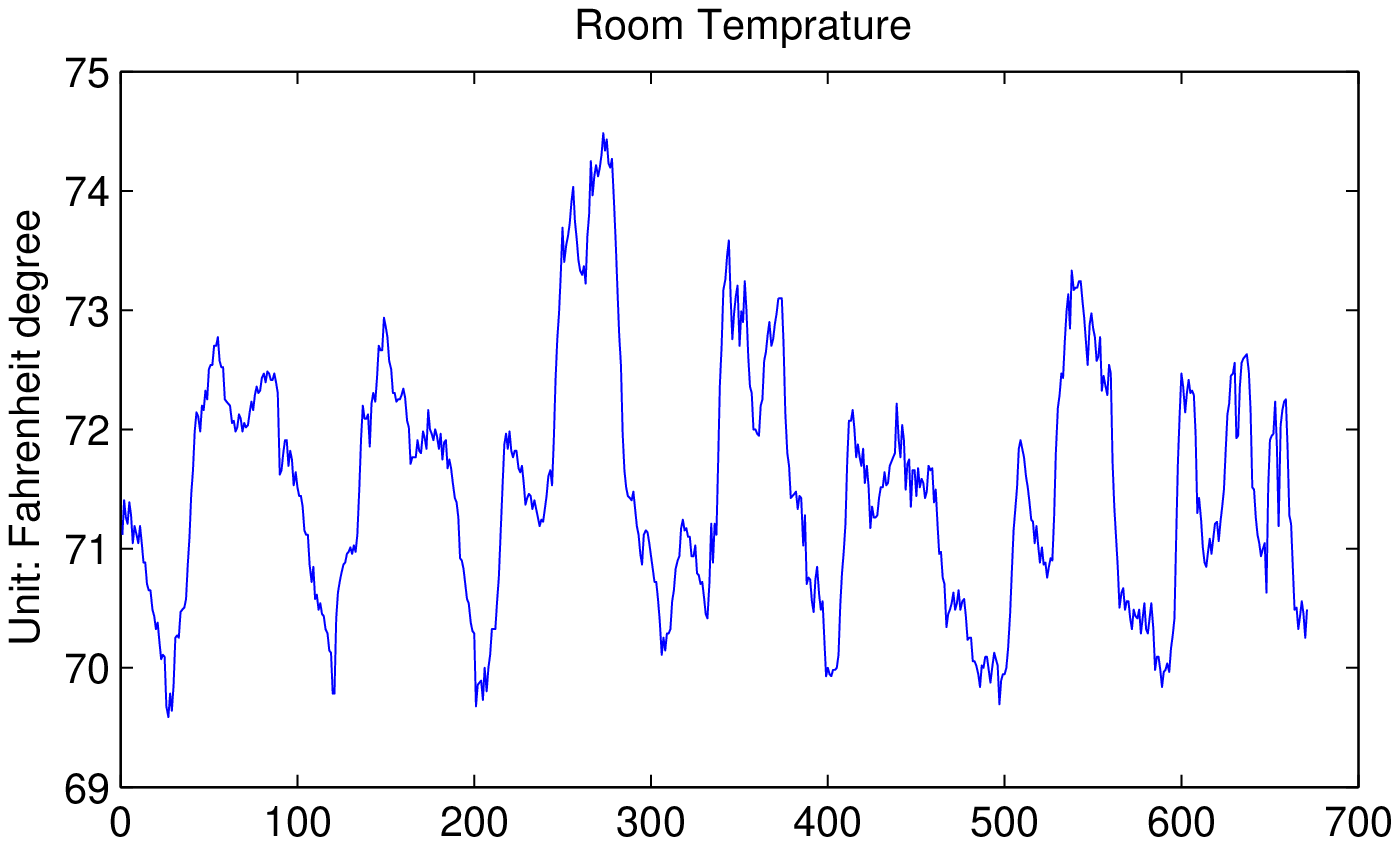}
                \caption{Room Temperature}
  \end{subfigure}
  \begin{subfigure}{0.32\textwidth}
                \centering
    \includegraphics[width=\textwidth]{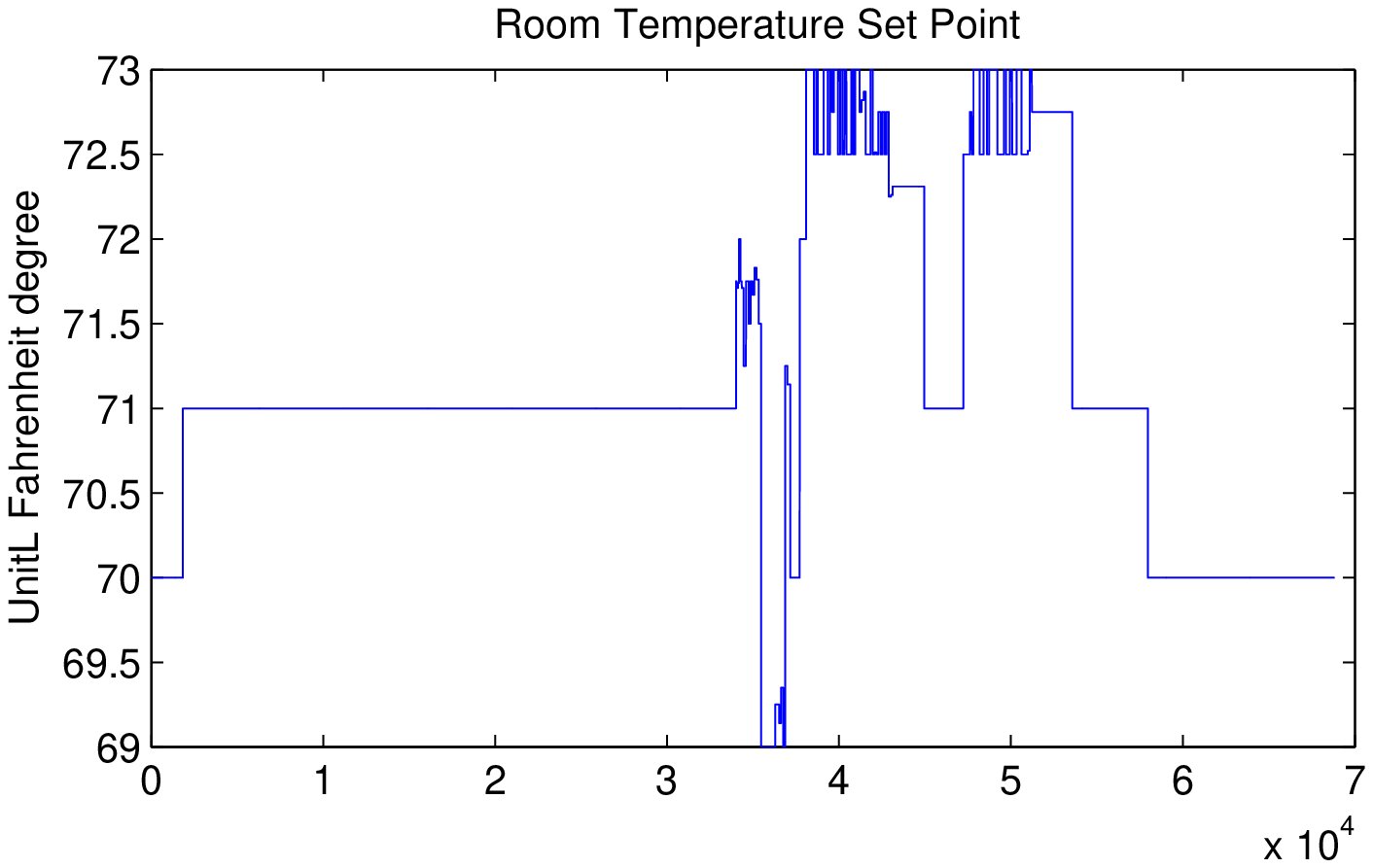}
                \caption{Room Temperature Set Point}
  \end{subfigure}
  \begin{subfigure}{0.32\textwidth}
                \centering
    \includegraphics[width=\textwidth]{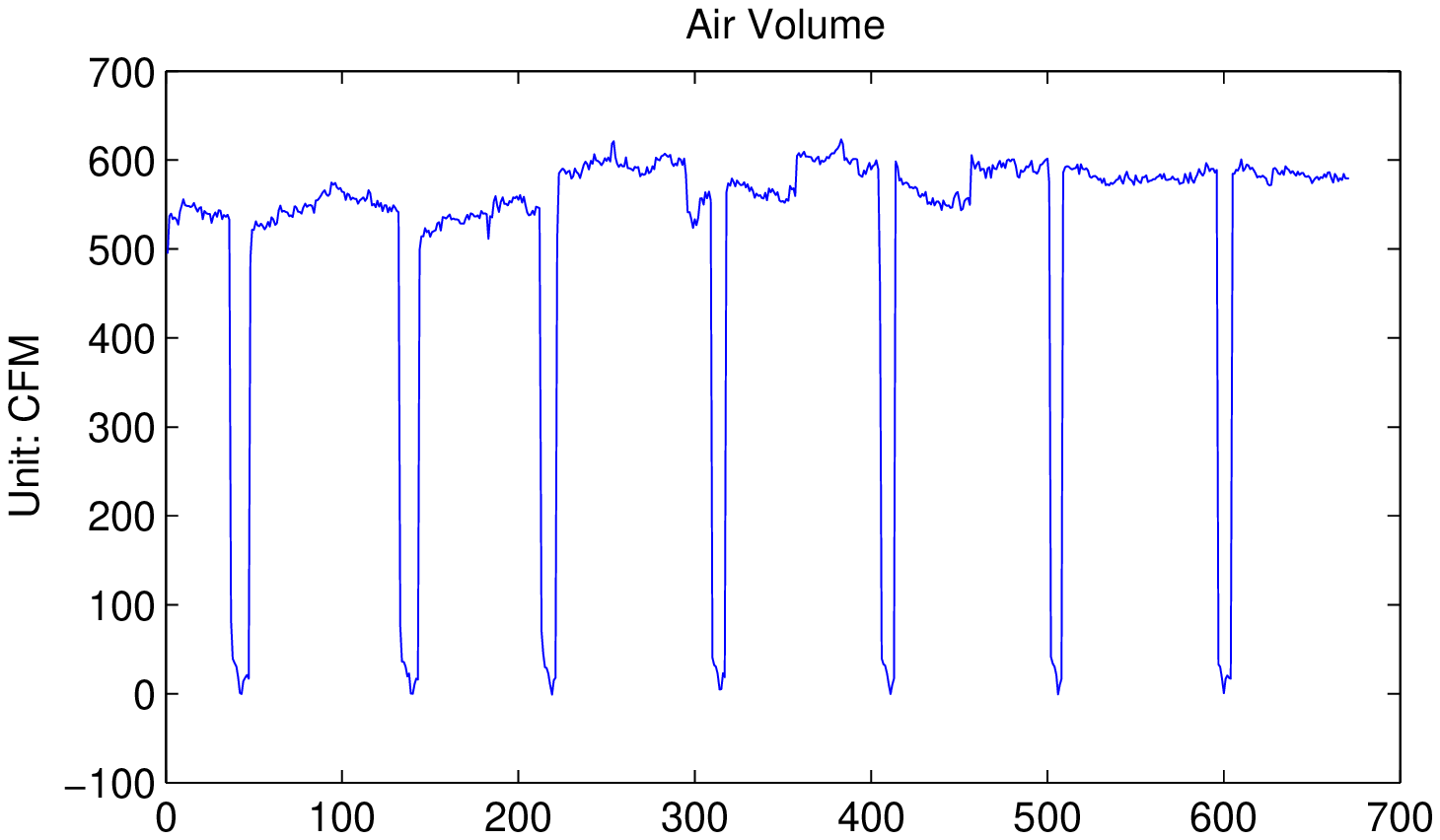}
                \caption{VAV Air Volume}
  \end{subfigure}
  \begin{subfigure}{0.32\textwidth}
                \centering
    \includegraphics[width=\textwidth]{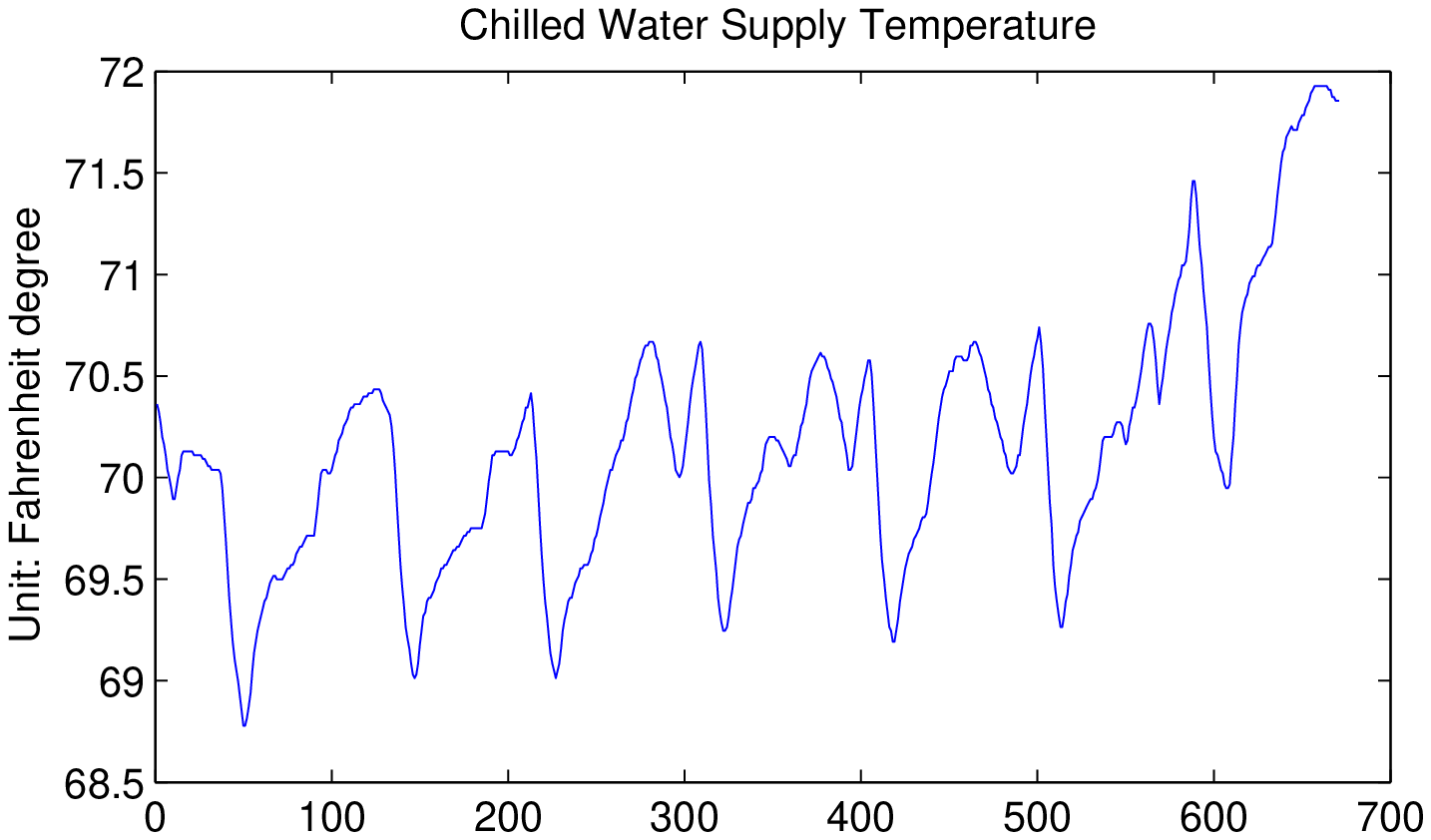}
                \caption{Chilled Water Supply Temperature}
  \end{subfigure}
\caption{Different types of sensors occupy different amplitude bins in the time domain with different short term dynamics.}
\label{fig:example}
\end{figure*}

\section{Methodology}
In this section, we describe the design and construction of the feature-vector we use to characterize sensor \emph{type}.  We
explain what it captures, fundamentally and, hence, why it works so well for building sensor data. Then, we discuss
the classification technique we apply and give a detailed description of the training and testing process. Finally, we articulate
a solution for identifying potentially misclassified streams, when no type-label ground truth is available.

\subsection{Feature Extraction}
Raw sensor time series\footnote{In this paper, we use the term ``trace'', ``readings'' and ``time series'' \textit{interchangeably}.} usually contain millions of readings which are too general to be useful for type classification.   We need to distill the information embedded in the reading patterns.
A signal in the time domain trends the amplitude of a sensor reading and different types of sensor generally occupy distinct
amplitude bins, as demonstrated in Figure~\ref{fig:example}. We can characterize the amplitude distribution of a signal in the time
domain by using the percentiles of the value distribution.
To identify outliers in the distribution, we pick the 50th percentile value (also known as the median) as a discriminator, which
is more robust to outliers skew than the average. 

\begin{figure}[ht!]
\centering
  \begin{subfigure}{0.22\textwidth}
                \centering
    \includegraphics[width=\textwidth]{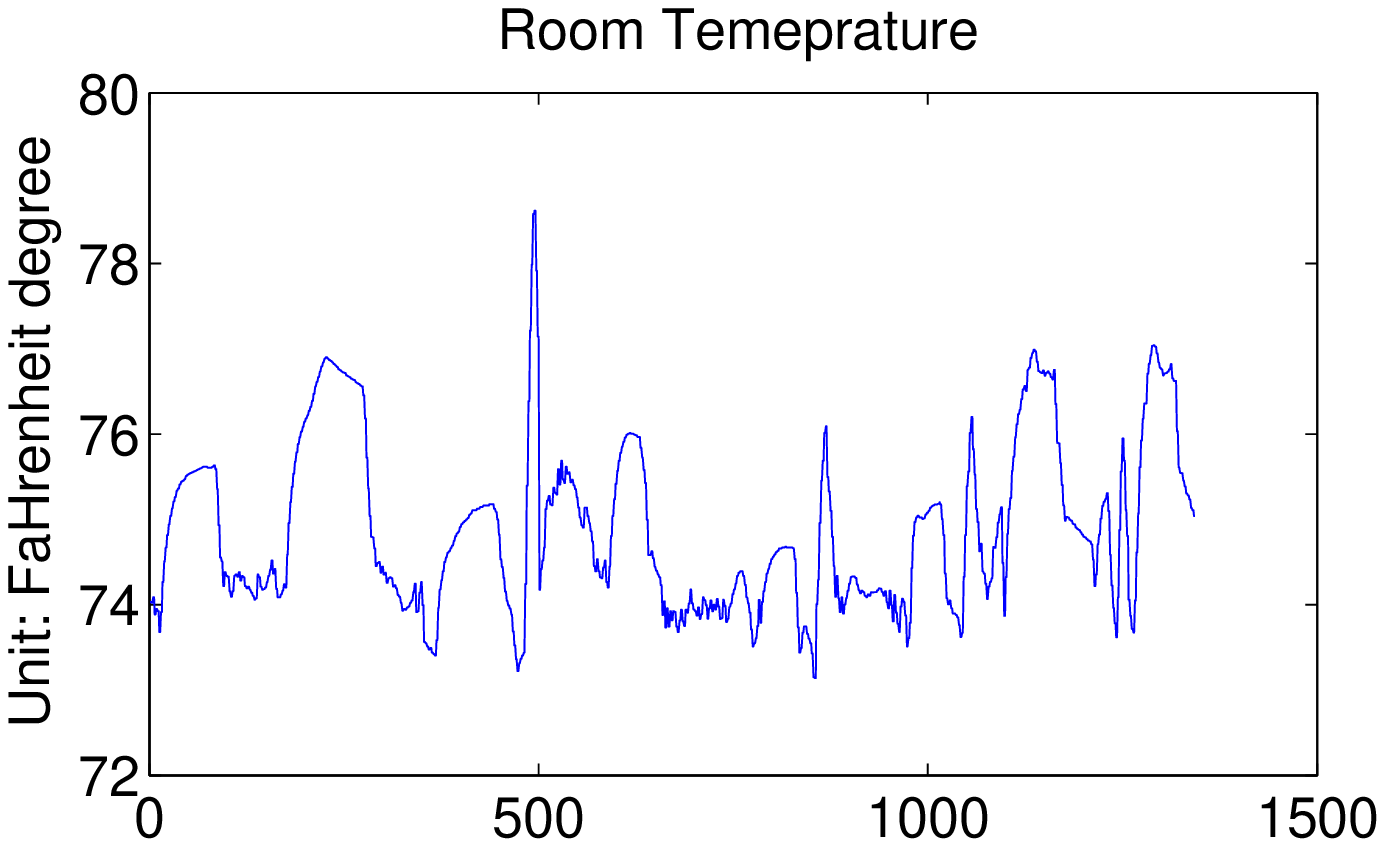}
                \caption{Room Temperature}
  \end{subfigure}
  \begin{subfigure}{0.22\textwidth}
                \centering
    \includegraphics[width=\textwidth]{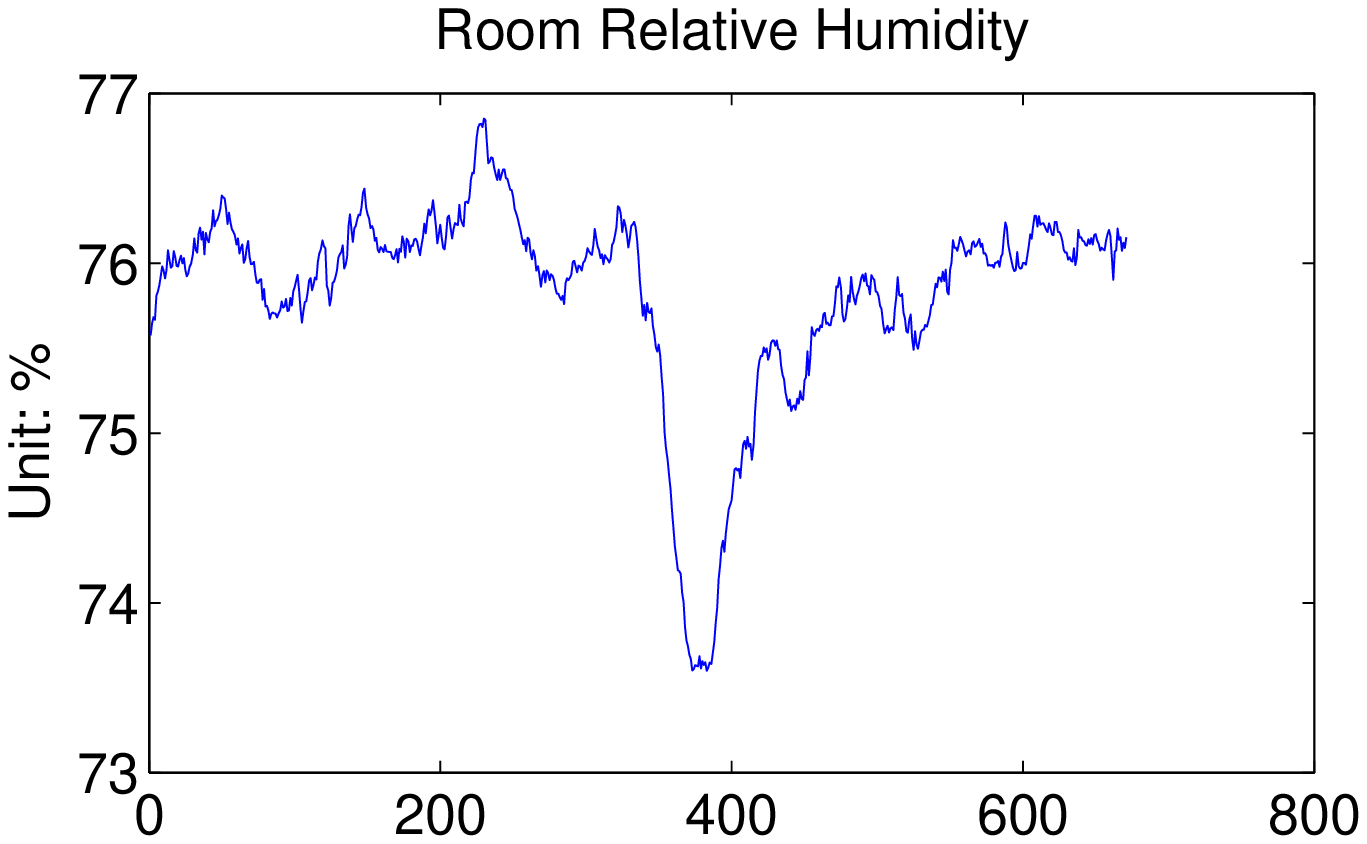}
                \caption{Humidity}
  \end{subfigure}
\caption{An example of two different types of sensors occupying the same amplitude bin.}
\label{fig:same_bin}
\end{figure}

Naturally, sensor reading value-ranges may overlap. For example, during a rainy season, the humidity in an
office can reach the range of 60-70 (percent) which is the same as typical temperature sensor readings (Fahrenheit).
If you do not consider measurement units, the distributions for each of the two types look similar. 
Simply relying on percentiles is not sufficient for differentiating sensor types. Figure~\ref{fig:same_bin} demonstrates this
point. To capture their difference we need to include the variance of the signal in our feature-vector.

When we extract features from a raw sensor readings, the original trace can span hours, days, or weeks, and
the trend can vary significantly, even from hour to hour. Extracting certain features, such as percentiles, and
variances over the entire sensor trace might miss  short term dynamics thus missing discriminating characteristics.
In contrast, computing features over short time windows can produce too much noise.  Too many  feature variables typically 
degrades classifier performance.
To succinctly summarize the dynamics of sensor traces, we apply feature extraction to every time window of fixed length on 
the original trace and compute the statistics of the accumulated features from windowed slices as the final feature set. 

Upon close inspection of the traces, we notice that the short-term dynamics of the phenomena being measured, could be used
to differentiate them.  Therefore, the distribution of short-term summary statistics can be used discriminate
traces by type.  We construct our feature vector as follows: first,
each single sensor signal is segmented into N
non-overlapping 45-minute long windows (we will discuss the decision of window length in later section). Second, within
each time window, we compute the median and variance of the signal, producing a vector of medians and a vector of
variances after the window slides over an entire trace: 
\begin{displaymath}
\begin{split}
MED = \{median^{1}, median^{2}, ..., median^{N}\}\\
VAR = \{variance^{1}, variance^{2}, ..., variance^{N}\}
\end{split}
\end{displaymath}
Where N is the number of time windows. The vector $MED$ and $VAR$ reflect short term changes but not all the intermediate values are useful for classification. Finally, we compute a statistical summary of the two vectors.   For each vector we compute the minimum, maximum, median and variance, resulting in a feature vector with eight variables:
\begin{displaymath}
\begin{split}
F = \{min(MED), max(MED), median(MED), var(MED),\\
 min(VAR), max(VAR), median(VAR), var(VAR)\}
\end{split}
\end{displaymath}
And $F$ is the feature vector for each sensor trace used in our classification process.

\subsection{Classification}
In general, ensemble learning methods obtain better predictive performance than
 any of the constituent learning methods as discussed in~\cite{ensem1,ensem2}, if the following assumptions hold~\cite{ensem3}: 1) the 
 probability of a correct classification by each individual classifier is greater than 0.5 and 2) the errors
  in predictions of the base-level classifiers are independent. Random forests~\cite{RF} have been widely used 
  and outperform a single tree classifier. They are also faster~\cite{cvpr} in training and testing compared to traditional classifiers such as SVM. The notion of randomized trees was first introduced in~\cite{RT} and further well developed in~\cite{RF}.

Random forests construct a multiple of classification trees. To classify an unlabeled object, we construct the feature vector and 
``feed'' the vector down each of the trees in the forest. Each tree gives a classification and we use it as a 
``vote" for that class. The forest chooses the class having the most votes over all the trees in the forest. 
The process proceeds as follows:
\begin{enumerate}
\item Sample N instances at random with replacement\footnote{An element may appear multiple times in the sample set.}, from the original data set. These samples will be the training set for growing this particular tree.
\item Specify M feature variables at random out of the total feature vector when growing each node of a tree. And the best split (measured by the information gain) on these M is used to split the node. The value of M is constant during the forest growing.
\item Each tree is grown to the largest extent possible without pruning.
\end{enumerate}
The randomness of this ensemble learning method occurs in the first two steps.  We set N equal the number of instances in the original training set, M equal the square root of the
number of original feature variables, and the number of the trees in the forest be 50. Usually these parameters are optimized
through cross-validation and we refer interested readers to~\cite{RF} for further deduction and proof of random forest.

All the instances in our data set are labeled with ground truth class. To train a random forest, we split the original set into two subsets, one for building 
a forest and one for testing the accuracy of the classifier. After the forest is built from training set, we learn the posterior probabilities of each class $c$ 
at each leaf node $l$ for each tree $t$: suppose that $T$ is the set of all trees, $C$ is the set of all classes and $L$ is the set of all leaves for a given 
tree $t\in T$. In the training stage the posterior probabilities $P_{t,l}(Y(i) = c))$ for each class $c\in C$ at each leaf $l\in L$, are learned for each tree $t\in T$. 
These probabilities are calculated as \textit{the ratio of the number of instance $i$ of class $c$ that reach $l$ to the total number of instances that reach $l$}. $Y(i)$ 
is the class label for instance $i$. To classify an instance in the testing stage, the feature vector of an instance is passed down each tree until reaching a leaf 
node, which gives a probability distribution. All the posterior probabilities accumulated from each tree are then averaged and the $argmax$ is taken as the class 
of the instance. Note that instead of letting each tree vote for on class as described in the original paper~\cite{RF}, we combine the results from classifiers for 
an instance by averaging their probabilistic predictions, in order to facilitate our technique used to help identify potential misclassified instances as described in the following section.  


\subsection{Quantify Classification Uncertainty}
Being able to measure the confidence of prediction results and to identify potential misclassifications, is vital to a learning process. 
It is trivial to 
identify misclassification when ground truth is available, but in many real-world cases ground truth is unavailable. 
Quantifying classification uncertainty can help identify potential misclassified instances and presents an opportunity to solicit 
the user for feedback that we can use to improve our results. To quantify the uncertainty of classifications in our learning process, we use
 the posterior probabilities learned in the random forest.

With the learned posterior probabilities for each class, at each leaf in each tree, we can compute the average probabilities 
for each class as follows:
\begin{displaymath}
    \bar P(Y(i)=c) = \frac{\sum_{t} P_{t,l}(Y(i)=c)}{|T^{'}|}, t\in T^{'}
\end{displaymath}
Where $T^{'}$ is the collection of trees in the forest where $P_{t,l}(Y(i)=c)\neq 0$, and $|\cdot|$ denotes the cardinality of a set. Given these averaged 
probabilities for each class, the forest produces a vector of class probabilities for each new instance as:
\begin{displaymath}
\textbf{Pr} = \{\bar P(Y(i)=c)\}, c\in C
\end{displaymath}
Suppose we have a probability vector $Pr_{1} = \{0.9, 0, 0, 0, 0, 0.1\}$ for instance $i_{1}$ and another vector $Pr_{2} = \{0.3, 0.25, 0.1, 0, 0.15, 0.2\}$ for instance $i_{2}$.
Both $i_{1}$ and $i_{2}$ will be assigned to the same class according to the class probability distribution, but the assignment of $i_{2}$ is less confident compared to that 
of $i_{1}$ because its predicted class probability has a less ``concentrated'' distribution. To measure the degree of ``uncertainty'' in classification of one instance, 
we compute the entropy of its class probability yielded by the forest. We rank the classification results and filter out the instances for 
further manual inspection whose entropy are above a threshold.  Inspection can help eliminate misclassifications.

\section{Evaluation}
To demonstrate the effectiveness of our methodology, we evaluate our classification technique in two different scenarios: a) intra-building, that is, the 
training and testing data for classification is taken from the same building, and b) inter-building, where the training and testing instances are from two 
distinct buildings. We also discuss how the amount of training instances and the window size of segment affect the performance of classification.
At last, we analyze the results of our solution for identifying potential misclassifications.

\subsection{Taxonomy}
\begin{figure}[ht!]
\centering
\includegraphics[width=0.4\textwidth]{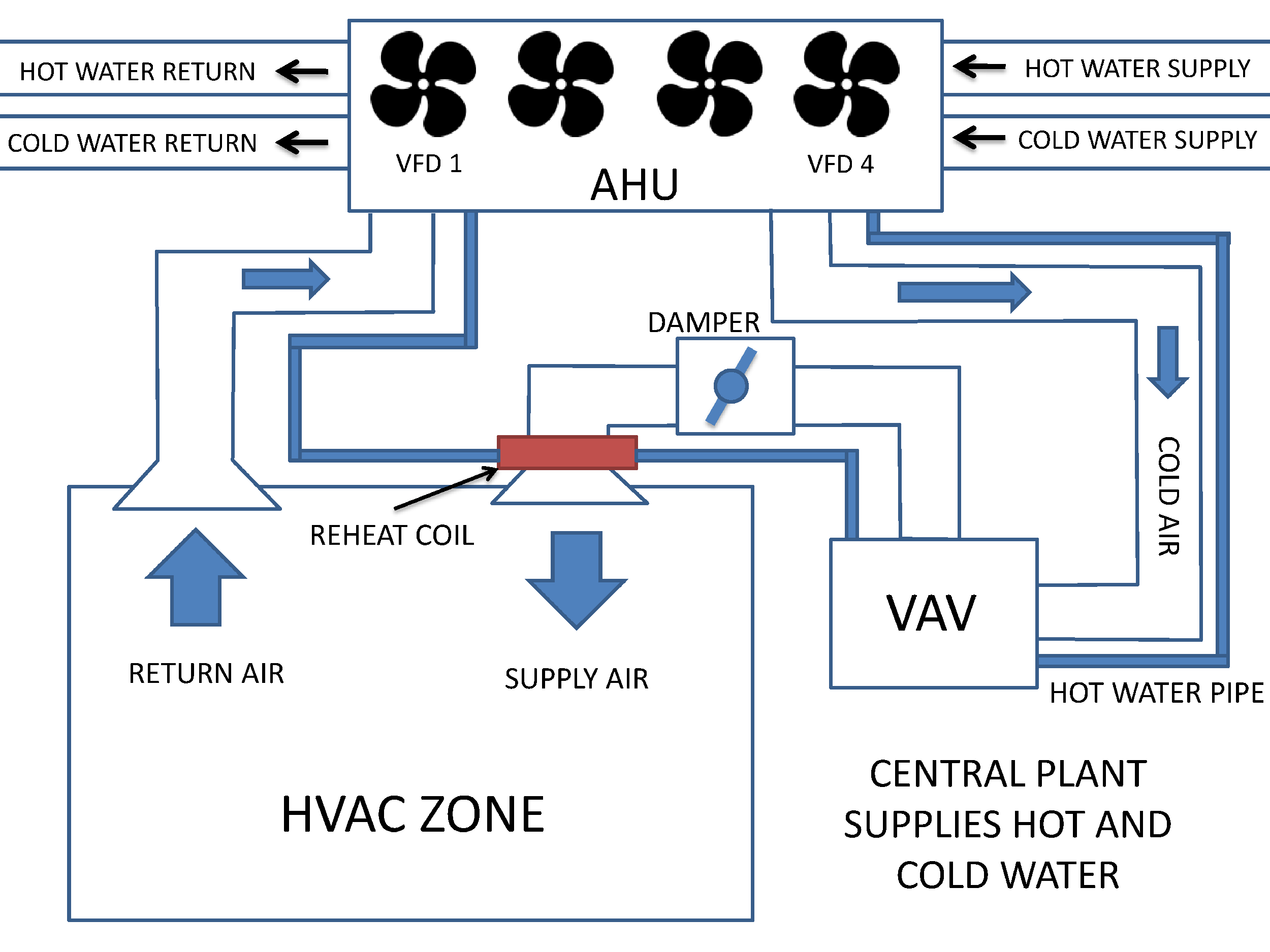}
\caption{A typical HVAC system consisting of water-based heating/cooling pipes and air circulation ducts.}
\label{fig:hvac}
\end{figure}

Most of the sensing points in the building can be classified into 6 general types, which we use in our work.
In this paper, we consider 6 types of sensors: $CO_{2}$, humidity, room temperature, setpoint, air flow volume, other temperature.  Room 
temperature includes only sensors that measure the air temperature of rooms (as ``room temp'' in Table~\ref{table:num}) and other temperature (as ``other temp'' 
in Table~\ref{table:num}) incorporates all other temperature measurements involved in an air ventilation system (illustrated in Figure\footnote{Included 
with permission from the authors of~\cite{sentinel}.}~\ref{fig:hvac}) such as supply air/return air/mixed air temperature and chilled or hot supply water/return water temperature. 
For set points, we assign only one general type which includes all set points for every actuator configured in the building.

\subsection{Experimental Setup}
We collected a week's worth of data from two separate buildings on two campuses. One is from the Rice Hall at the University of Virginia, where 
the sense points report to a database~\cite{trane} anywhere between every 10 seconds to every 10 minutes. The other building is the Sutardja Dai Hall (SDH) at UC Berkeley, 
where the deployed sensors~\cite{keti, bacnet} transmit to an archiver~\cite{smap} periodically from anywhere between every 5 seconds to every 10 minutes. The number for each type 
of sensor in each building is summarized in Table~\ref{table:num} and the type \emph{ground truth} for each sense point is manually expanded based on the metadata in the database.

\begin{table}[ht!]
\centering 
\begin{tabular}{c c c}
\hline 
Type & Rice & SDH \\ 
\hline\hline 
$CO_{2}$ & 16 & 52 \\ 
Humidity & 48 & 52 \\
Room temp & 142 & 216 \\
Setpoint & 265 & 819 \\
Air volume & 12 & 158 \\ 
Other temp & 119 & 37 \\ \hline
Sum & 602 & 1334 \\ \hline
\end{tabular}
\caption{Number of Sensors by Type}
\label{table:num} 
\end{table}

All of our learning and classification processes are implemented based on the scikit-learn~\cite{scikit} library, which is an open-source machine learning package 
implemented mostly in Python providing a rich set of APIs.

\subsection{Baseline and Metrics}
\label{sec:baseline}
As a baseline to compare our proposed approach against, we adopt a simple feature extraction scheme for each trace $F=\{med, var\}$, where $med$ and $var$ is simply 
the $median$ and $variance$ computed over the entire trace.

For classification, we measure the averaged cross-validation accuracy in two different scenarios (intra- and inter- buildings). In the intra-building case, the 
data from a single building is split into training and testing sets, where the results illustrate how accurately the type information can be inferred using local 
within-building information. For inter-building case, the experiment performs training and testing across buildings, i.e, train the classifier on the data from building A 
and test it on building B.  
This set of experiments tests how well we can apply the classification boundaries from one building and apply it to another.

For identifying potential misclassifications, we choose the true-positive rate (TPR, also known as recall), false-positive rate (FPR, also known as fall-out) and positive predictive 
value (PPV, also known as precision) as metrics to evaluate the performance of our entropy-based approach when making different choice of threshold. Particularly, under our misclassification
identification context, a true-positive (TP) is when an instance considered to be misclassified is really misclassified while a false-positive (FP) is when an instance considered to be misclassified 
is instead a correct classification.

\subsection{Classification Accuracy}
We run the two sets of experiments described above, i.e, the intra- and inter- building tests, to examine the effectiveness of feature design and measure how well 
the classifier performs. The classification results are summarized from Table~\ref{table:rice}-\ref{table:sdh_x}. 
In the table, each row is specific to a \emph{type} and each column is the \emph{percentage} of the full data set that was used for training.
Each cell shows two values.  The value without parentheses is the average classification accuracy for the richer feature-vector. 
The value in parentheses is the average classification accuracy for the approach described in 
Section~\ref{sec:baseline}. These are compared throughout the table.
The last column summarizes ``leave-one-out'' cross-validation\footnote{In LOO cross validation, each training set takes all the instances except one with the test set being 
the sample held out.} results for each approach.

\subsubsection{Intra Building Performance}
From the last column in Table~\ref{table:rice} and~\ref{table:sdh}, we see that type classification in a single building achieves accuracy of $\sim$92\% and $\sim$98\% on Rice Hall 
and SDH respectively, for leave-one-out (LOO) cross validation. The accuracy for the baseline is also shown in the table (in parentheses). The only type we have difficulty differentiating is ``other temp'', which includes 
temperature measurements for air and water in the ventilation system, and particularly, the return air temperature measurements (as illustrated in Figure~\ref{fig:hvac}) are 
almost identical to the ones measuring air temperature in rooms because what the return duct exhausts is mostly the air from a room.

\begin{table*}[ht!]
\centering 
\begin{tabular}{c | c | c | c | c | c | c}
\hline 
Type & 5\% & 10\% & 20\% & 33\% & 50\% & LOO\\ 
\hline\hline 
$CO_{2}$ & 51.3 (60.7) & 83.7 (94.0) & 98.4 (98.5) & 100.0 (100.0) & 93.8 (100.0) & 93.8 (100.0)\\ \hline
Humidity & 59.6 (61.8) & 66.8 (67.9) & 80.8 (77.1) & 82.3 (74.3) & 87.5 (83.8) & 83.3 (89.8)\\ \hline
Room temp & 89.0 (88.5) & 93.0 (94.0) & 95.6 (94.7) & 93.3 (95.9) & 97.2 (93.7) & 95.1 (95.6)\\ \hline
Setpoint & 97.0 (93.1) & 97.5 (95.2) & 99.2 (95.7) & 99.2 (96.5) & 98.5 (97.4) & 99.2 (97.8)\\ \hline
Air volume & 22.2 (21.8) & 35.5 (30.5) & 46.7 (49.6) & 79.2 (66.7) & 41.7 (83.3) & 83.3 (75.0)\\ \hline
Other temp & 54.7 (47.2) & 64.8 (59.0) & 70.0 (71.0) & 72.7 (75.4) & 74.7 (71.7) & 74.8 (83.1)\\ \hline
Overall & 80.4 (78.3) & 85.9 (84.4) & 90.0 (88.3) & 90.9 (90.0) & 91.4 (90.2) & 91.7 (93.3)\\ \hline
\end{tabular}
\caption{Intra-building Classification Accuracy for Rice Hall}
\label{table:rice} 
\end{table*}

\begin{table*}[ht!]
\centering 
\begin{tabular}{c | c | c | c | c | c | c}
\hline 
Type & 5\% & 10\% & 20\% & 33\% & 50\% & LOO\\ 
\hline\hline 
$CO_{2}$ & 80.4 (63.4) & 87.8 (75.8) & 91.4 (74.6) & 89.5 (83.7) & 92.3 (86.5) & 96.2 (76.9)\\ \hline
Humidity & 91.6 (92.4) & 94.4 (92.1) & 97.6 (95.2) & 98.1 (98.1) & 100.0 (100.0) & 98.1 (98.1)\\ \hline
Room temp & 98.3 (96.0) & 98.9 (96.3) & 99.2 (95.6) & 98.4 (94.7) & 97.7 (93.1) & 99.1 (95.8)\\ \hline
Setpoint & 99.2 (89.0) & 99.6 (90.4) & 99.5 (91.4) & 99.7 (91.8) & 99.5 (90.7) & 99.5 (93.3)\\ \hline
Air volume & 78.4 (41.8) & 87.1 (47.4) & 92.7 (52.9) & 96.8 (57.1) & 98.7 (55.3) & 97.5 (57.1)\\ \hline
Other temp & 23.7 (19.4) & 38.4 (28.7) & 62.3 (36.5) & 68.9 (48.7) & 75.7 (59.9) & 73.0 (59.5)\\ \hline
Overall & 93.4 (81.4) & 95.6 (83.7) & 97.2 (85.2) & 97.8 (86.6) & 98.2 (86.0) & 98.3 (87.7)\\ \hline
\end{tabular}
\caption{Intra-building Classification Accuracy for SDH}
\label{table:sdh} 
\end{table*}

\subsubsection{Inter Building Performance}
This set of experiments illustrates how accurately we can learn the type information of one building based on the knowledge from another building. The overall classification accuracy 
achieved for the two buildings by training on the entire data set (train on SDH for Rice and train on Rice for SDH) is around 82\%, as seen from the last columns in Table~\ref{table:rice_x} 
and~\ref{table:sdh_x}. Particularly, we see that the accuracy for ``other temperature'' in Rice is abnormal compared to the rest of the
results. The issue with classifying ``other temperature'' in Rice is that there are many sensing points measuring the temperature of supply and return cold/hot water utilized in the 
ventilation system, which are absent in the Berkeley building as a training set. Therefore the feature of these traces cannot be learned from SDH and causes problems in classifying these traces.

\begin{table*}[ht!]
\centering 
\begin{tabular}{c | c | c | c | c | c | c}
\hline 
Type & 5\% & 10\% & 20\% & 33\% & 50\% & 100\% \\ 
\hline\hline 
$CO_{2}$ & 29.7 (44.4) & 45.6 (56.9) & 75.0 (75.0) & 93.8 (72.9) & 93.8 (75.0) & 87.5 (93.8)\\ \hline
Humidity & 50.9 (30.5) & 72.1 (26.7) & 76.2 (28.6) & 76.4 (21.1) & 89.6 (16.3) & 87.5 (26.5)\\ \hline
Room temp & 97.6 (92.7) & 99.4 (91.4) & 100.0 (90.9) & 97.2 (92.5) & 100.0 (93.1) & 100.0 (91.8)\\ \hline
Setpoint & 97.8 (94.7) & 98.2 (94.6) & 98.0 (92.8) & 97.7 (91.4) & 97.5 (92.3) & 98.9 (92.6)\\ \hline
Air volume & 57.5 (21.2) & 58.3 (18.3) & 66.7 (23.3) & 63.9 (25.0) & 70.8 (20.8) & 83.3 (25.0)\\ \hline
Other temp & 5.3 (5.8) & 10.8 (6.2) & 11.1 (6.9) & 16.8 (10.2) & 18.9 (10.5) & 19.3 (12.9)\\ \hline
Overall & 73.1 (69.1) & 76.9 (68.7) & 78.3 (68.7) & 79.1 (68.5) & 81.3 (68.7) & 81.9 (70.3)\\ \hline
\end{tabular}
\caption{Inter-building Classification Accuracy for Rice Hall}
\label{table:rice_x} 
\end{table*}

\begin{table*}[ht!]
\centering 
\begin{tabular}{c | c | c | c | c | c | c} \hline
Type & 5\% & 10\% & 20\% & 33\% & 50\% & 100\%\\ 
\hline\hline 
$CO_{2}$ & 63.5 (94.2) & 96.9 (96.2) & 90.8 (96.9) & 93.6 (94.9) & 92.3 (95.2) & 98.1 (98.1)\\ \hline
Humidity & 67.4 (28.8) & 86.3 (47.4) & 98.1 (45.0) & 96.2 (41.0) & 98.1 (25) & 98.1 (44.2)\\ \hline
Room temp & 78.0 (78.0) & 78.2 (75.8) & 72.9 (73.8) & 77.9 (76.7) & 80.3 (77.3) & 53.2 (77.3)\\ \hline
Setpoint & 77.4 (53.3) & 83.3 (50.8) & 86.5 (53.4) & 87.9 (54.4) & 87.2 (62.0) & 91.8 (58.1)\\ \hline
Air volume & 13.8 (34.7) & 15.2 (33.1) & 37.8 (25.1) & 42.4 (32.9) & 50.3 (30.3) & 71.5 (38.8)\\ \hline
Other temp & 48.3 (51.4) & 49.7 (53.1) & 58.4 (4.6) & 45.0 (52.3) & 45.9 (54.1) & 67.6 (51.4)\\ \hline
Overall & 68.2 (55.5) & 74.1 (54.3) & 78.4 (54.5) & 80.3 (56.2) & 81.2 (60.1) & 83.0 (59.6)\\ \hline
\end{tabular}
\caption{Inter-building Classification Accuracy for SDH}
\caption*{Each table shows the averaged classification accuracy of experiments where different percentage of the complete set is used as training set (denoted as `X\%'). In the percentage analysis, each percentage is repeated 1/percentage times and the averaged accuracy is presented. LOO cross validation accuracy is also shown for the intra-building test case. On average, our solution outperforms the baseline approach (shown in parentheses).}
\label{table:sdh_x} 
\end{table*}

\subsection{Learning Bootstrapping}
We also experiment with different amount of training instances to examine how that affects the classification accuracy, which gives some insight on how many instances are needed 
to bootstrap the classification process. In Table~\ref{table:rice}-~\ref{table:sdh_x}, 
the last columns demonstrate how accurately we can do 
classification on average. There also remains the question of how many number of instances we need to bootstrap the learning process in both of the intra- and inter- building cases. 
To examine the impact of number of instances on classification accuracy, we use different percentage of the original data set as training set, i.e, 5\%, 10\%, 20\%, 33\%, 50\%, and the 
results are presented in the first five columns in each table. For each percentage of training instances used, we apply stratified sampling\footnote{The sampled set contains the same 
percentage of samples of each class as the original complete set.} on the original set and the remaining instances are used as testing set. We repeat the same percentage 1/percentage times 
to reduce random errors and get an averaged accuracy for that percentage. We can clearly see a trend that more training instances yield better classification results in all cases. 
However, we can also notice that after the training set includes about 20\% of the complete set (which is $\sim$120 instances and $\sim$260 instances for Rice and SDH respectively) the 
accuracy doesn't increase too much even reaching 100\% of the complete set. This indicates that we don't need too many instances to bootstrap the learning process within or across buildings 
to accomplish sensor type classification tasks.

\subsection{Window Size Sensitivity}
\begin{figure}[ht!]
\centering
	\includegraphics[width=0.45\textwidth]{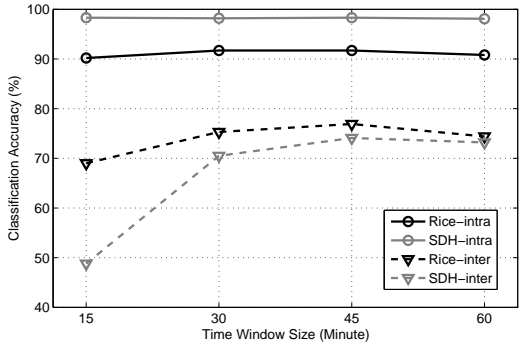}
\caption{Classification accuracy of intra- and inter- building cases against different size of time window: a window size of 45 minutes is optimal for our classification tasks.}
\label{fig:window}
\end{figure}

All the classification results, thus far, were obtained using features extracted in 45-minute window slices on the original sensor traces. We study how different windows sizes affect the classification performance. Figure~\ref{fig:window} shows those results. The \emph{intra} case performs LOO cross validation while \emph{inter}
case runs 10-fold cross validation. For the intra-building case, the classification is not sensitive to different window sizes as seen in the figure: basically, accuracy stays almost the same 
for both buildings because within the same building, as long as we can capture the short term characteristics of sensor dynamics in the windowed time slots, the size of the time window 
doesn't make too much difference. However, for the inter-building case, the time window size matters in the way that usually the local micro-climate in one building can be quite different 
from another, we need to ``tune'' this common short term window to capture the dynamics that can be used to learn type-related information across buildings. Therefore, in order to achieve 
decent type classification accuracy across buildings, (i.e, use information from one building to help classify the traces in another building), we still need to optimize the size of time window, 
which is 45 minutes in our case. This ``tuning'' is significant for the learning process across buildings and is straightforward to perform.

\subsection{Identifying Potential Misclassifications}
\begin{figure*}[ht!]
\centering
	\begin{subfigure}{0.48\textwidth}
                \centering
		\includegraphics[width=\textwidth]{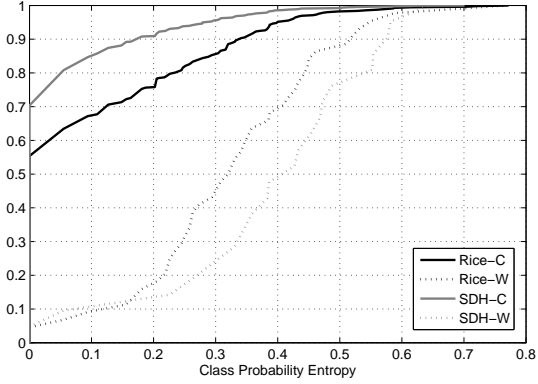}
                \caption{Intra-building Classification}
                \label{fig:cdf_intra}
	\end{subfigure}
	\begin{subfigure}{0.48\textwidth}
                \centering
		\includegraphics[width=\textwidth]{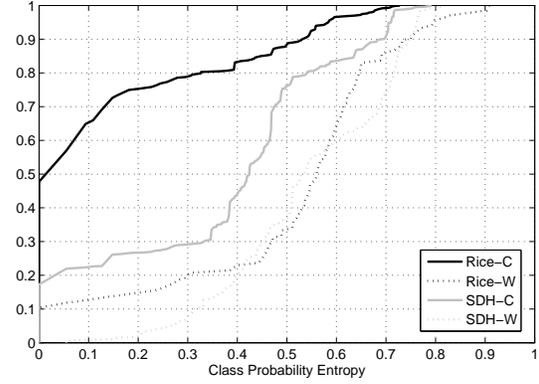}
                \caption{Inter-building Classification}
                \label{fig:cdf_inter}
	\end{subfigure}
\caption{The CDF curves depict the class probability entropy distribution for the collection of correct (`-C' in solid lines) and wrong (`-W' in dotted lines) classifications in the intra- and inter- building test cases. The collection of correct classifications has a distinct distribution from the collection of wrong ones.}
\label{fig:cdf}
\end{figure*}

\begin{figure*}[ht!]
\centering
	\begin{subfigure}{0.48\textwidth}
                \centering
		\includegraphics[width=\textwidth]{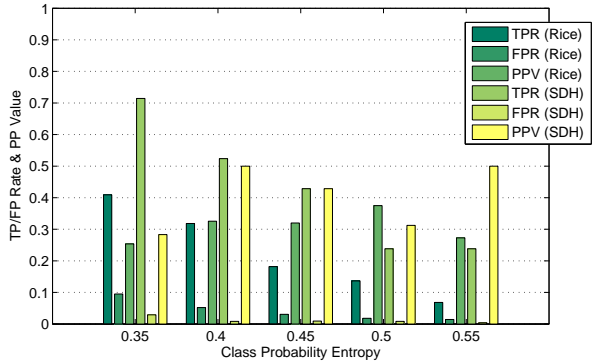}
                \caption{Intra-building Performance}
                \label{fig:roc_intra}
	\end{subfigure}
	\begin{subfigure}{0.48\textwidth}
                \centering
		\includegraphics[width=\textwidth]{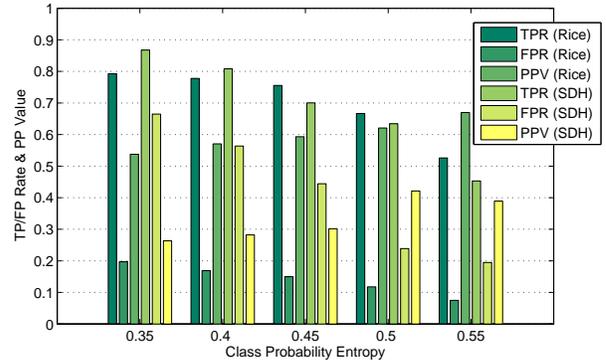}
                \caption{Inter-building Performance}
                \label{fig:roc_inter}
	\end{subfigure}
\caption{The ROC curves depict the sensitivity of misclassification identification to different entropy threshold value. Choosing a threshold somewhere between 0.4 and 0.45 achieves the best compromise between recall and precision. }
\label{fig:roc}
\end{figure*}

As we discussed in early section, being able to quantify the confidence in classifications and identify misclassified instances in our sensor type classification is vital to improving 
the overall accuracy considering that in many cases our technique is used there will be the absence of ground truth. As an intermediate step to identify potentially misclassified instances, 
we propose to quantify the ``uncertainty'' of classification with an entropy-based approach described in Section 3.3. Figure~\ref{fig:cdf} shows the CDF of class probability entropy of 
classification in the intra- and inter- building scenarios. We see that the collection of correct classifications (in solid lines) has a distinct distribution from the collection of 
misclassification (in dotted lines). Based on such distinction in the distribution, we can choose a certain entropy value as a threshold and filter out all the classified instances whose 
class probability entropy are greater than the threshold outputted by the forest. Figure~\ref{fig:roc} gives a summary of the performance of our entropy-based approach to 
identifying potential misclassification. Here are some definitions needed to understand the statistics:

$S_{1}$: the set of instance whose class probability entropy is greater than the threshold.

$S_{2}$: the set of instance falling in $S_{1}$ that is misclassified in the classification process.

$S_{3}$: the set of instance falling in $S_{1}$ that is correctly classified in the classification process.

$S_{4}$: the set of instance that is misclassified in the classification process.

$S_{5}$: the set of instance that is correctly classified in the classification process.

And the TPR, FPR and PPV are defined as: 
\begin{displaymath}
TPR = \frac{|S_{2}|}{|S_{4}|},\quad
FPR = \frac{|S_{3}|}{|S_{5}|},\quad
PPV = \frac{|S_{2}|}{|S_{1}|},\quad
\end{displaymath}
Where $|\cdot|$ is the cardinality of a set. We see that as the threshold value increases, both of the TPR (recall) and FPR (fall-out) decrease while the PPV (precision) keeps increasing. 

In our case, a smaller threshold essentially leads to a larger population of instances being filtered out as potential misclassification ``candidates'', which helps identify more real 
misclassified instances. However, the more candidates we filter our, the more instances we need to manually inspect, which inevitably leads to a lower precision of the identification process. 
So we want to strike a balance between achieving a high recall rate as well as maintaining a high precision. As a result, based on the observation from Figure~\ref{fig:roc}, we suggest 
picking a threshold value somewhere between 0.4 and 0.45 is appropriate. To note, we have 50 and 13 misclassified instances for Rice and SDH respectively for the intra-building testing case. 
In the intra-building case, such a threshold (0.4-0.45) helps identify $\sim$30\% of the misclassified instances for Rice and $\sim$50\% for SDH while resulting in that $\sim$70\% and $\sim$50\% 
of the instances being manual inspected are actually correct classifications, for Rice and SDH respectively. As for the inter-building case, our approach is able to identify $\sim$75\% of the 
misclassified instances for both Rice and SDH with an overhead of $\sim$40\% and $\sim$70\% in the candidate inspection, for Rice and SDH respectively.


\section{Discussion}
There are several aspects of our work that we left out or did not have time to explore more deeply.
First we go over the expansion of \emph{type} classes and how we could increase coverage of sensor types in future work. 
We discuss how we could improve classification accuracy
by looking for data sources outside the building data sets. We also discuss why principal component analysis is an aspect that we
did not explore in depth and how the principal components can change from building to building.  Finally, we 
explain how our misclassification identifier could be used to improve classification results.

\begin{table*}
    \centering 
    \begin{tabular}{c|c|c|c}
        \hline 
        Building & Set of Best Features & Acc. on All & Acc. on Best Set \\ 
        \hline\hline 
        Rice & min(MED), med(MED), med(VAR), var(VAR) & 88.7\% & 91.5\% \\ \hline
        SDH & min(MED), max(MED), max(VAR), med(VAR) & 97.1\% & 97.8\% \\\hline
    \end{tabular}
    \caption{Classification accuracy on all the features and on the best set of features in intra-building test for each building: the best feature sets are obtained by exhausting all the feature combinations and running on a single decision tree with leave-one-out cross validation. The best feature set is different for each building.}
    \label{table:feature} 
\end{table*}

\subsection{Extension of Taxonomy and Class Scope}
Our taxonomy covers 5 specific and one general sensor \emph{type}. We could extend the class scope to include more sensor types and make our technique more versatile. There are many types of sensors in modern buildings and the sensing fabric in smart buildings continue to diversify, e.g., occupancy sensors, light sensors, etc. We also want to build a deeper taxonomy for certain types.  
For instance, there are set points for very different actuators.  Temperature set-points drive the HVAC system, while the air quality set-point drives the filters and air mixers. Being able to differentiate between these can help enable general control applications in buildings.

\subsection{Improvement on Classification Accuracy}
The learning and classification processes in our work relies only on a set of general features. However, we wish to explore how using external or domain-specific knowledge could help improve the classification accuracy. For instance, if we know the humidity in rooms will increase due to a rain forecast, then we could search for traces with increases in average in reading values as external knowledge to help identify humidity traces. 

\subsection{Feature Importance and Selection}
In our study, we did not delve into the the importance of features (i.e. principle component analysis) 
because the feature vector contains only eight variables.  Therefore, doing classification in a hyperspace of only eight dimensions 
is not computationally expensive -- even if some of the vector elements carry redundant information. More importantly, 
that selecting the set of principle features for each building results in using a different feature set 
(as demonstrated in Table~\ref{table:feature}) per building.  This makes classification across buildings impossible. Still, 
evaluating the principle components and uncovering overlap is
important for obtaining optimal classification performance for intra-building tasks and single-type analysis.

\subsection{Reducing Misclassification Iteratively}
In cases where no ground truth labels are available, an entropy-based approach can be used in an iterative manner to improve classification results. 
In each iteration, 
only a few examples (on the top of the entropy-based ``uncertainty'' ranking list) are inspected and corrected, and the corrected instances could be
 added to the training set.  The training 
and classification process is repeated until some criteria is satisfied. We expect the number of examples needed for manual inspection 
will be dramatically reduced in each iteration and overall, compared to a one-time inspection of candidates filtered by some threshold value.
Such an interactive, supervised learning process can produce better classification results and reduce the human labeling effort needed.

\section{Related Work}
To the best of our knowledge, we are the first to approach the problem of sensor-type classification of physical data in buildings.
  We describe the closest, related work in different problem domains and describe work that uses the random forest as a tool.

There has been much research work on type classification in the context of audio~\cite{audio1,audio2}, music~\cite{music1,music2}, video~\cite{video1,video2}, 
web query~\cite{query1,query2} and human activity~\cite{activity1, activity2}. The goal of~\cite{audio1} is to classify audios into categories such as speech, 
music, background sound and silence using support vector machines, and the work in~\cite{audio2} addresses the same problem with a HMM-based statistical model. 
Examples of music genre (i.e, jazz, pop and so forth) classification are~\cite{music1,music2}, which use GMM with EM algorithm and logistic regression
respectively. And commonly used features for these audio-related classification work are MFCC, zero crossing rate, energy/power and spectral/temporal statistics. 
For video type classification, texture and color-based features are used to classify videos into classes including cartoon, commercial, news and so on 
with decision tree~\cite{video1} and neural network~\cite{video2}. Query categorization has also been researched, ~\cite{query1} exploits a rule-based classifier
while~\cite{query2} uses a Markov random walk model. There is also work on human activity classification in general cases~\cite{activity1} (i.e, running, walking 
and sitting) and home setting~\cite{activity2} (i.e, sleeping, toiletting and showering) using accelerometer data with voting-based classifier and HMM with 
conditional random fields respectively. In contrast, our work is focused on sensor type classification using ensemble learning technique.

Random forests have been applied in many different areas~\cite{RF1,RF2,RF3,RF4}. ~\cite{RF1} uses a gene as a feature 
to classify microarray data. ~\cite{RF2} uses the intensity of hundreds of measured metabolites from medical subjects as features, to classify the 
subjects into groups of normal, diseased and diseased with drug treatment. Random forests have also been used in~\cite{RF3} to classify 
 objects in images with image-relevant features. In the area of remote sensing, ~\cite{RF4} utilizes user-defined parameters as features 
 to classify land cover 
types. In our work, we use simple and general statistical feature-set for type classification.

There is work leveraging percentile-based features in time series data for different classification purposes. 
Tarzia~\cite{ABS} 
et. al use a certain percentile in the audio spectrum to classify the current room location. Wang~\cite{business} et. al utilize percentile-based features in audio to characterize occupancy and noise levels. For comparison, we use percentile statistics in sensor time series as 
part of our feature-set to differentiate between different sensor types in commercial buildings.

\section{Conclusion}
We describe a general, simple yet effective feature extraction design in support of sensor type classification with time series data. By experimenting 
with over 2000 streams from two buildings on two campuses, our technique, which leverages an ensemble learning method, is able to achieve an accuracy more than 
92\% and 82\% for testing within building and across buildings, respectively. We also discuss that how to choose the window size applied 
to a slice of the original time series and how the number of training instances affects classification accuracy. In general, around 100 instances are
enough to bootstrap the learning process in the case of 6 types of sensors. Another important contribution of our paper is a probability-based solution for identifying 
potentially misclassified instances. With the use of probabilities produced by the random forest, in both of the intra- and inter- building learning cases, we are able 
to identify at least 30\% of the misclassifications.

Our technique can act as a tool for metadata construction for building sensors. For cases where type information of sensors is missing, our technique can 
help infer and generate the type metadata. In cases where metadata is available in an inconsistent manner within/across buildings, our solution 
can be used to verify type information and unify the naming schema across platforms in different buildings. Questions remain about how broadly we 
can expand our taxonomy and further study the scalability of our technique.


\bibliographystyle{abbrv}
\bibliography{buildsys}

\end{document}